\documentclass[conference]{IEEEtran}
\IEEEoverridecommandlockouts
\usepackage{cite}
\usepackage{amsmath,amssymb,amsfonts}
\usepackage{algorithmic}
\usepackage{algorithm2e}
\usepackage{graphicx}
\usepackage{textcomp}
\usepackage{xcolor}
\usepackage{hyperref}
\usepackage{makecell}

\usepackage{caption} 
\RestyleAlgo{ruled}

\makeatletter
\newcommand{\linebreakand}{%
  \end{@IEEEauthorhalign}
  \hfill\mbox{}\par
  \mbox{}\hfill\begin{@IEEEauthorhalign}
}
\makeatother

\def\BibTeX{{\rm B\kern-.05em{\sc i\kern-.025em b}\kern-.08em
    T\kern-.1667em\lower.7ex\hbox{E}\kern-.125emX}}
\begin{document}

\title{Online Transfer Learning for RSV Case Detection\\

}

\author{\IEEEauthorblockN{Yiming Sun}
\IEEEauthorblockA{\textit{Department of Electrical and} \\\textit{ Computer Engineering} \\
\textit{University of Pittsburgh}\\
Pittsburgh, USA\\
yis108@pitt.edu}
\and
\IEEEauthorblockN{Yuhe Gao}
\IEEEauthorblockA{\textit{Department of} \\\textit{ Biomedical Informatics} \\
\textit{University of Pittsburgh}\\
Pittsburgh, USA\\
yug51@pitt.edu}
\and
\IEEEauthorblockN{Runxue Bao}
\IEEEauthorblockA{\textit{Department of Electrical and} \\\textit{ Computer Engineering} \\
\textit{University of Pittsburgh}\\
Pittsburgh, USA\\
runxue.bao@pitt.edu}
\and
\IEEEauthorblockN{Gregory F. Cooper}
\IEEEauthorblockA{\textit{Department of} \\\textit{Biomedical Informatics} \\
\textit{University of Pittsburgh}\\
Pittsburgh, USA\\
gfc@pitt.edu}
\linebreakand 
\IEEEauthorblockN{Jessi Espino}
\IEEEauthorblockA{\textit{Department of Biomedical Informatics} \\
\textit{University of Pittsburgh}\\
Pittsburgh, USA\\
jue@pitt.edu}
\and 
\IEEEauthorblockN{Harry Hochheiser}
\IEEEauthorblockA{\textit{Department of Biomedical Informatics} \\
\textit{University of Pittsburgh}\\
Pittsburgh, USA\\
harryh@pitt.edu}
\and 
\IEEEauthorblockN{Marian G. Michaels}
\IEEEauthorblockA{\textit{Department of Pediatrics} \\
\textit{University of Pittsburgh}\\
Pittsburgh, USA\\
michmg@pitt.edu}
\linebreakand 
\IEEEauthorblockN{John M. Aronis}
\IEEEauthorblockA{\textit{Department of Biomedical Informatics} \\
\textit{University of Pittsburgh}\\
Pittsburgh, USA\\
jma18@pitt.edu}
\and
\IEEEauthorblockN{Chenxi Song}
\IEEEauthorblockA{\textit{Department of Biomedical Informatics} \\
\textit{University of Pittsburgh}\\
Pittsburgh, USA\\
chs342@pitt.edu}
\and
\IEEEauthorblockN{Ye Ye$^\dagger$\thanks{$\dagger$ Corresponding author}}
\IEEEauthorblockA{\textit{Department of Biomedical Informatics} \\
\textit{University of Pittsburgh}\\
Pittsburgh, USA\\
yey5@pitt.edu}}



\maketitle

\begin{abstract}
Transfer learning has become a pivotal technique in machine learning and has proven to be effective in various real-world applications. However, utilizing this technique for classification tasks with sequential data often faces challenges, primarily attributed to the scarcity of class labels. To address this challenge, we introduce \textit{Multi-Source Adaptive Weighting} (MSAW), an online multi-source transfer learning method. MSAW integrates a dynamic weighting mechanism into an ensemble framework, enabling automatic adjustment of weights based on the relevance and contribution of each source (representing historical knowledge) and target model (learning from newly acquired data). We demonstrate the effectiveness of MSAW by applying it to detect Respiratory Syncytial Virus cases within Emergency Department visits, utilizing multiple years of electronic health records from the University of Pittsburgh Medical Center. Our method demonstrates performance improvements over many baselines, including refining pre-trained models with online learning as well as three static weighting approaches, showing MSAW's capacity to integrate historical knowledge with progressively accumulated new data. This study indicates the potential of online transfer learning in healthcare, particularly for developing machine learning models that dynamically adapt to evolving situations where new data is incrementally accumulated.

\end{abstract}


\section{Introduction}

Machine learning has made substantial advancements in recent decades, with its applications spanning a wide range of fields such as computer vision (CV), natural language processing (NLP), speech recognition, and autonomous driving. Despite these achievements, machine learning in biomedicine faces significant challenges, particularly in data collection. The acquisition of labeled data can be very costly or even infeasible due to factors like ethical considerations, patient privacy, and the scarcity of certain diseases. These challenges have led researchers to increasingly rely on utilizing data from related domains that have a more abundant supply of data.


To better leverage knowledge from these related yet distinct domains, transfer learning has emerged as a promising approach to enhance model performance. Fundamental to transfer learning are the concepts of the `source domain' and `target domain'. The source domain, which is usually rich in data, serves as the foundation for knowledge extraction. It is characterized by its abundance of well-labeled datasets, which aid models in understanding and encoding complex patterns. In contrast, the target domain, where this pre-learned knowledge is used, often features a more limited dataset. Transfer learning bridges this gap by adapting models trained on the source domain to function effectively in the target domain, thereby helping address the challenges associated with data scarcity.


Traditional transfer learning approaches \cite{SHDA-RF, KPDA, SHFR} typically depend on having complete datasets from both the source and target domains at the start of the training process. However, this is often not feasible, especially in dynamic environments where target domain data are collected progressively. To tackle this issue, the concept of online transfer learning \cite{zhao2014online, kang2020online, zhao2010otl} has been introduced. This method is designed for scenarios where target domain data are acquired incrementally over time, while source domain data are already accessible before training begins. In such cases, transferring knowledge from the source domain becomes crucial, particularly because the limited initial data in the target domain may be insufficient for effective learning. The extensive and diverse information available from the source domains can significantly compensate for this shortfall, providing an initial knowledge base that the model can build upon as more target domain data becomes available. Therefore, the efficiency and effectiveness of learning in the target domain are greatly enhanced by the transferred knowledge from the source domains.


Online transfer learning entails leveraging knowledge from a static source domain and applying it to an ongoing, evolving target task. The model is incrementally updated as new target data are received. A crucial element of this process is the utilization of a loss function to fine-tune the model in accordance with the latest incoming data, thus ensuring its continued relevance and efficiency. This adaptability renders online transfer learning particularly suitable for real-time applications, effectively handling issues such as concept drift \cite{agrahari2022concept} and providing a more versatile approach to knowledge transfer.


In traditional transfer learning, the distinction between various domains is clear and easily identifiable. This clarity allows for straightforward categorization based on common characteristics, such as grouping images by distinguishing features like those of cats versus dogs.  However, online transfer learning presents a nuanced challenge as it focuses on the evolution of a single task over time, making the determination of where to apply temporal segmentation less clear. This difficulty arises because the task's concept remains unchanged, while the data representing the task may significantly change due to temporal factors. This evolution can alter the relevance of past information to current or future tasks, making it difficult to understand how these varying time periods connect and influence one another. Unlike static categorizations, where similarities and differences are often clear and measurable, the dynamic nature of temporal domains means that the relevance of past data to the present task can be fluid and elusive.
For instance, in healthcare, temporal segmentation might involve dividing patient data into annual periods to capture seasonal disease trends. With multiple source domains, integrating them optimally into an ensemble model is challenging. A critical question arises: how do we determine which source domains are most relevant to the current period (or target domain) and which are less relevant or unrelated? 



Our proposed method employs a sequential data acquisition strategy to tackle this challenge. Consider a healthcare scenario where the task is predicting patient outcomes in a new medical facility. To address initial data scarcity in this
facility, our approach significantly relies on pre-trained models derived from source domain data, such as historical patient
outcomes from established facilities. These pre-trained source models form the initialization of our ensemble model. This ensemble is adept at autonomously evaluating and determining
the relevance of each source model’s learned patterns to the
current patient data in the new facility (target model). For instance, a source
model trained on a large urban hospital might be initially
prioritized for its broad spectrum of cases. As more patient
data accumulates in the new facility, the ensemble dynamically
adjusts its reliance on this urban model, potentially shifting
focus to another source model that aligns more closely with emerging local trends. This strategic utilization of pre-trained
source models ensure that the most effective ones are prioritized during the critical early stages of training in the target
domain.


As the target domain accumulates more data, the ensemble model modifies its approach. Initially dependent on pre-trained source models, it gradually begins to prioritize the continually updated target model. This change is twofold: a shift in reliance from source to target models and a nuanced adjustment in the weight each model holds in decision-making. This transition enables the model to evolve alongside the growing target domain dataset. The distinct roles of shifting focus and adjusting weights demonstrate the adaptability of our online transfer learning approach, particularly in environments where data are collected over time.

Our framework makes two main contributions. First, we introduce an online multi-source transfer learning architecture for machine learning using data over multiple years. Second, we explore the performance of different ways of generating weights for the ensemble model to improve its accuracy. 

We evaluate the effectiveness of our approach using real-world Respiratory Syncytial Virus (RSV) patient encounter data that were collected over multiple years from Emergency Departments (EDs) at University of Pittsburgh Medical Center (UPMC). Our primary target-domain goal is to utilize patients' signs and symptoms to accurately classify them as either having RSV or not, a critical step towards improving infectious disease case detection using EHRs. RSV is a major cause of respiratory infections that can lead to hospitalizations \cite{piedimonte2014respiratory}. The burden of RSV continues to grow, especially in young children aged less than 5 years old\cite{shi2017global, burdenRSV}. RSV also has been identified as an important disease causing hospitalizations of the elderly and high-risk adults \cite{RSVelderly}. The disease typically exhibits a seasonal pattern in case numbers \cite{griffiths2017respiratory}. In this study, we segment the RSV case data into distinct domains based on a defined annual cycle, with each domain spanning from June of the previous year to May of the current year. This segmentation strategy is crucial to our framework, aiming to improve RSV case detection accuracy by using data from multiple preceding seasons as source inputs. For instance, we use patient encounter data from June 2011 to May 2018 to develop an ensemble model for classifying encounters from June 2018 to May 2019. The experimental results indicate that our approach improves classification performance, particularly related to public health planning and resource allocation for anticipated RSV seasons. 


\section{Related Work}


Traditional transfer learning methodologies typically assume that the source and target domains are predefined, aiming to transfer knowledge from the source to the target domain. Many comprehensive surveys \cite{zhuang2020comprehensive, weiss2016survey, zhang2022transfer, agarwal2021survey, bao2023survey} have detailed and categorized a broad range of transfer learning approaches, evaluating factors such as algorithm types and application scenarios.
A notable recent development is the growing body of literature reviews focusing on the integration of transfer learning with other advanced machine learning techniques. These include combinations with deep learning \cite{tan2018survey, iman2023review}, reinforcement learning \cite{liang2019survey, zhu2023reinforcement}, and federated learning \cite{saha2021federated, hallaji2022federated}.
Additionally, some reviews have specifically targeted the application of transfer learning in key areas such as CV \cite{wang2018deep, zhang2024trade}, NLP \cite{alyafeai2020survey, liu2019survey, ruder2019transfer}, healthcare \cite{yu2022transfer, sufian2020survey, wei2023transfer, wei2023unfairness, lai2024language}, and wireless communication \cite{wong2022transfer, maray2023transfer}, providing insights into the practical implications and utility of transfer learning in these specific fields.

Online transfer learning has emerged to address situations where data are received sequentially and are not available at the start of training. This approach is particularly relevant in dynamic environments where real-time data processing is crucial. Some popular online transfer learning methods \cite{wu2019online,lai2024adaptive, hu2023leveraging, yan2017online} have demonstrated excellent performance in many real-world scenarios. Online transfer learning methods can be categorized into data-based methods and model-based methods. 

Data-based transfer learning methods involve the transfer of original data to a target domain, allowing the target models to be trained with these augmented data, thereby enriching the available data within the target domain. Online Heterogeneous Knowledge Transition\cite{wu2019online} trains the target model using data that co-occur in source and target domains. The co-occurrence data can be a set of text-image pairs, for example, the image includes a red car and the text is `red; wheel; transportation'.

Conversely, model-based transfer learning methods focus on firstly building models within the source domain, then adapting both the model structure and parameters for the target domain. Here, `model structure' refers to the architecture or configuration of the model (such as layers in a neural network), while `model parameters' are the specific coefficients that are learned from the source data. 
The target model leverages insights from the source domain to potentially improve performance. Ensemble models, which combine the outputs or features of multiple models, are commonly employed in model-based methods to further improve accuracy and robustness \cite{lai2024adaptive, hu2023leveraging}.
Online Heterogeneous Transfer by Hedge Ensemble \cite{yan2017online} handles the online heterogeneous transfer problem by updating weights of offline and online classifiers based on the Hedge weighting strategy. 





For our online transfer learning algorithm, we used RSV detection as a representative example task. 
The use of machine learning for RSV detection has been growing. Tso et al. \cite{tso2022machine} developed an extreme gradient boost decision tree model to detect RSV cases in both emergency department and inpatient settings, using clinical measurements from EHRs, including respiratory rate and peripheral oxygen saturation. 
In another study, Cord et al. \cite{cord2023machine} assessed different methodologies such as Logistic Regression, Gaussian Naive Bayes, and Random Forest, ultimately choosing the Light Gradient Boosting Machine algorithm
for their RSV detection models. 

\section{Methodology}

In this section, we begin by outlining the problem formulation for the online multiple source-domain scenario. Following this, we delve into the base model, setting the stage for the introduction of our proposed method.

\subsection{Problem Formulation}


We address the challenge of binary classification in environments where data are dynamic, arriving sequentially and often associated with specific time frames such as epidemiological seasons. Our focus is on creating a method to analyze information collected across seasons, using Bayesian networks as the basis of our classifier training. In practical scenarios, data from each season can be viewed as a separate domain. This temporal aspect presents a distinct challenge, especially at the beginning of a new season when data are scarce. To tackle this problem, we suggest a transfer learning strategy where the current season is the target domain, and data from several past seasons are used as source domains. In transfer learning, research \cite{ji2023prediction} has indicated that simply combining multiple source domains into one is inferior to employing a multi-source strategy.
Thus, our method uses data from multiple source domains, represented as $D^S = \{D_S^1, D_S^2, \ldots, 
D_S^n\}$, where each $D_S^i$ is data from a specific season, together with a single target domain $D_T$. We leverage historical data from past seasons, assuming that patterns and knowledge can effectively inform and improve the classification model for the current season.

\subsection{Base Model}

Our experiment study begins with the implementation of the Naive Bayes classifiers. 
The effectiveness of Naive Bayes stems from its simplicity, which facilitates tracking model changes across seasons over time, and its computational efficiency, particularly when the features are relatively independent given the class label. An additional benefit of the Naive Bayes model is its statistical efficiency, which allows one to construct a model from a small number of instances, making it particularly well-suited for online learning with only target data.

\subsection{Online Multi-source Ensemble Model}

\begin{figure}[h]
    \centering
    \includegraphics[width=0.48\textwidth]{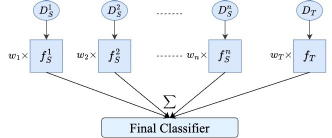}
    \caption{Architecture of the proposed ensemble model.}
    \label{fig:model}
\end{figure}

In our framework, for each source domain $D_S^i$, we train an offline base model $f_S^i$ as a classifier on this domain. Similarly, a specific online training classifier $f_T$ is designated for the target domain. The architecture of our proposed model is depicted in Fig. \ref{fig:model}. Central to our approach is an ensemble model, formulated as:
\begin{equation}
    f = w_1 f_S^1 + \ldots + w_n f_S^n + w_T f_T
\end{equation}
Here, $w_i$ represents the weight parameter for the $i$-th source domain classifier, and $w_T$ denotes the weight parameter for the target domain classifier. Each source domain classifier $f_S^i$ is pre-trained on its respective domain data, while the target domain classifier $f_T$ is dynamically trained and updated with new instances from the target domain as they become available.

Our ensemble method capitalizes on the strengths of individual classifiers trained on distinct historical data patterns from each source domain, integrating these with the versatile classifier of the target domain. This integration aims to enhance the model's overall classification capabilities and accuracy, which is especially beneficial when dealing with limited initial data in the target domain. For weight allocation in our ensemble model, adaptive weights can provide a dynamic adjustment mechanism.


\begin{algorithm}
\caption{Multi-Source Adaptive Weighting (MSAW)}
\SetAlgoLined
\textbf{Input}: Pre-trained source models $f_S^1(\cdot), \ldots, f_S^n(\cdot)$, Target model $f_T(\cdot)$, Target Dataset $X=\{X_j\}$ for $j=1,2,\ldots$, Weight penalty $\alpha \in \left(1,\infty \right)$, Target weight factor $\beta \in \left(0,1\right)$.

\textbf{Initialization}: Initialize source model weights $w_S^i = 1.0$ for $i = 1,2,\ldots, n$\, target model weight $w_T= 1.0$, and current data volume in target domain $j$. 

\While{receive new data point $X_j$}{
    Increase target data volume $j \leftarrow j+1$ \\
    Adjust target weight: $w_T \leftarrow w_T \times \beta \times j$\\
    Normalize weights: 

    \indent\hspace{0.5cm}$\displaystyle w_S^i \leftarrow \frac{w_S^i}{\sum_{i=1}^n w_S^i + w_T},$ $\displaystyle w_T \leftarrow\frac{w_T}{\sum_{i=1}^n w_S^i + w_T}$
    \\
    Get combined probability for final classification: \\ \indent\hspace{0.5cm}$prob = \sum_{i=1}^{n} w_S^i \times f_S^i(X_j)+w_T \times f_T(X_j)$
    
    \For{$i = 1$ to $n$}{
        \If{$f_S^i(X_j)$ classifies incorrectly}{
            Adjust weight: $w_S^i \leftarrow w_S^i \times \frac{\sqrt{j}}{\sqrt{j}+ \alpha} $
        }
    }
    \If{$f_T(X_j)$ classifies incorrectly}{
        Adjust weight: $w_T \leftarrow w_T \times \frac{\sqrt{j}}{\sqrt{j}+ \alpha}$
    }
    Update $f_T$ with $X_j$
    
    Return $prob$ as the classifying probability for $X_j$
}
\label{alg2}
\end{algorithm}

Our proposed Multi-Source Adaptive Weighting (MSAW) method, detailed in Algorithm \ref{alg2}, involves updating all weights, covering those of the target domain classifier and source domain classifiers. As new training instances are processed, the algorithm assesses the accuracy of each classifier, adjusting weight based on their classifying performance. Weights of accurate classifiers are unchanged while those of inaccurate classifiers are penalized, diminishing their impact on future classifications. MSAW also dynamically adjusts the weights of classifiers based on changes in data volume. 
As the amount of target domain data increases, the target classifier's weight is proportionally enhanced. This increase is quantified by multiplying the weight with a factor that signifies the quantity of target data, a process facilitated by the variable $j$. The variable $j$ serves as a sequential index for each new target data point, effectively tracking the expansion of the target dataset. As more target instances are received (i.e., as $j$ increases), the target classifier's weight is correspondingly adjusted upwards. This adjustment ensures that the weight of the target classifier is attuned to the scale of the available data, enabling the algorithm to progressively prioritize more recent and relevant insights. Simultaneously, the source domain classifiers' weights are adjusted primarily based on their classification performance. 
This dual mechanism—adapting weights for both classification accuracy and data volume—enables a more dynamic and responsive approach in the MSAW algorithm. The target classifier's weight becomes increasingly significant as more target data become available, ensuring that the model's classifications are continually refined and aligned with the latest trends and patterns in the target domain.



We hypothesize that the adaptive nature of the learnable weight assignment strategy in the proposed algorithm will outperform static weight assignment strategies. This hypothesis rests on the idea that dynamically adjusting weights enables the ensemble model to evolve in sync with incoming target data, potentially leading to more precise and adaptable classifications. The three static weighting assignment strategies for comparison are: (1) The `Equal Weights' method which assigns each source model a constant and equal weight. (2) 
The `Volume-based' method assigns weights to source models based on the size of their training datasets. (3) The `Time-based' method combines individual models based on their temporal proximity to the testing season (target domain), where `time distance' is defined as the number of seasons between a model's training data and the testing period.

\section{Dataset}

The research data were sourced from the University of Pittsburgh's Neptune Research Data Warehouse \cite{visweswaran2022atomic}, and the study protocol received approval from the University of Pittsburgh Institutional Review Board (Protocol STUDY20030193). The dataset consists of 2,263,360 encounters from five UPMC EDs (Presbyterian University Hospital, Children's Hospital, McKeesport Hospital, Mercy Hospital, and Shadyside Hospital) in Allegheny County over nine years from January 2011 to May 2020. In regions with temperate climates, RSV infections predominantly manifest during the colder winter months. The onset of these infections is usually observed around October and November in the Northern Hemisphere. Typically, the highest incidence of RSV cases, or the peak period, occurs in January and February. As the season progresses, the frequency of RSV infections starts to decline, with the final cases of the season generally being recorded between March and May \cite{griffiths2017respiratory}. Given the seasonal nature of RSV outbreaks, we defined one season as spanning from June of one year to May of the following year, rather than using a calendar year division for data segmentation.
Table \ref{tab:example} details the sizes of encounters for each season. An encounter was classified as related to RSV if it included one of any of the following International Classification of Diseases (ICD) codes at discharge; encounters that did not include any of these ICD codes were categorized as non-RSV. 

\begin{itemize}

\item 79.6	Respiratory syncytial virus
\item 466.11 Acute bronchiolitis due to respiratory syncytial virus
\item 480.1	Pneumonia due to respiratory syncytial virus
\item B97.4	Respiratory syncytial virus as the cause of diseases classified elsewhere
\item J12.1	Respiratory syncytial virus pneumonia
\item J20.5	Acute bronchitis due to respiratory syncytial virus
\item J21.0	Acute bronchiolitis due to respiratory syncytial virus
\end{itemize}

\begin{table}[htbp]
\renewcommand{\arraystretch}{1.3} 
\caption{Sizes of datasets}
\centering
\begin{tabular}{|c|c|c|}
\hline
\textbf{Season}& 
\makecell{\textbf{Number of RSV}\\ \textbf{Labeled Cases}}&
\makecell{\textbf{Number of Non-RSV}\\ \textbf{Labeled Cases}} \\ \hline
2011-2012 & 251 & 259,596 \\ \hline
2012-2013 & 238 & 258,024 \\ \hline
2013-2014 & 132 & 254,076 \\ \hline
2014-2015 & 179 & 250,476 \\ \hline
2015-2016 & 200 & 252,432 \\ \hline
2016-2017 & 381 & 257,172 \\ \hline
2017-2018 & 408 & 243,708 \\ \hline
2018-2019 & 507 & 240,312 \\ \hline
2019-2020 & 888 & 244,380 \\ \hline
\end{tabular}
\label{tab:example}
\end{table}

Our dataset contains 7,614 features, including demographic information, lab test results, and NLP-extracted clinical findings, detailed in Table \ref{tab:features}.
The clinical findings were obtained by using
the MetaMap parser \cite{aronson2010overview}, which applies pattern-matching and deduction rules. The clinical notes for each encounter serve as input for the parser, which extracts a set of clinical findings and their values. The values for each finding are either present (P), negated (N), or missing (M), with missing denoting that clinicians did not mention the finding in the note, as per the parser's output. The recommendation to designate missing values in NLP-retrieved findings from EHRs as a distinct category was made by \cite{pineda2015comparison}. In cases where an encounter had multiple clinical notes, we used the union of all clinical findings across all the notes. Contradictions across notes were resolved in favor of present findings, and this also applied to multiple occurrences of a finding in a single note.

\begin{table}[ht]
\renewcommand{\arraystretch}{1.3} 
\caption{Features in research datasets}
\centering
\begin{tabular}{|p{0.35\linewidth}|p{0.55\linewidth}|}
\hline
\textbf{Features} & \textbf{Distinct Values (Descriptions)}  \\ \hline
age & less5 ($<$5), ge5less18 (5-17), ge18less65 (8-64), ge65 (65+), M (value is missing)  \\ \hline
gender & female, male, unknown, M (value is missing)  \\ \hline
ethnicity & Hispanic or Latino, not Hispanic or Latino, not specified, M (value is missing)  \\ \hline
race & black, white, other, not specified, M (value is missing)  \\ \hline
7541 UMLS CUIs extracted from free-text reports by MetaMap & P (present), N (negated), and M (value is missing). One CUI example is C0859927 (Increased work of breathing).  \\ 
\hline
69 laboratory results & H (high), L (low), N (normal), M (value is missing). One lab example is oxygen saturation.  \\ 
\hline
\end{tabular}
\label{tab:features}
\end{table}


\section{Experiments \& Results}

\subsection{Experiment Settings}



We employed an online multi-source transfer learning ensemble framework to enhance model's adaptability in a specific target domain. This framework incorporated datasets from previous seasons, each treated as a distinct source domain. 
We evaluated models' performance using the Area Under the Receiver Operating Characteristic Curve (AUROC), an important measure for gauging a model's discriminative performance when classifying positive and negative cases. 


In the specified algorithm, two hyperparameters, $\alpha$ and $\beta$, play crucial roles in the dynamic adjustment of model weights during the learning process. The setting of $\alpha$ and $\beta$ is consistent across all domains. The hyperparameter $\alpha$ is set to $\log_2 10$ = 3.32, a value selected after experimenting with several hyperparameter settings. This value for $\alpha$ was chosen to balance the weight penalty applied to a classifier's weights upon making an incorrect classification. This particular value for $\alpha$ strikes a balance between being too lenient and too punitive, allowing for a gradual decrease in a classifier's influence while still retaining its contributions where they prove useful.

The hyperparameter $\beta$, set at $1/200000$, plays a critical role in the volume-adaptive component of our algorithm. This parameter is instrumental in fine-tuning the adjustment of the target weight $w_T$, aligning it with the growing volume of incoming data. Here, the factor $j$ serves as an index representing the sequence of data points being processed. The specific choice of $1/200000$ for $\beta$ strikes a delicate balance. When the number of instances has not yet reached 200,000, this configuration ensures that the target weights are subtly suppressed, preventing premature dominance in the model's decision-making process. However, once the count surpasses this threshold, the target weights are progressively amplified, reflecting the increased reliability and relevance of the more substantial data corpus. This ensures that the target model's influence is appropriately scaled, growing more pronounced as the volume of target data becomes more significant, thereby sensitively responding to the increase in data volume while simultaneously tempering the rate of weight adjustment. 

By adopting this value for $\beta$, the algorithm ensures a balanced and responsive adaptation to the accumulating target data. Such a setup allows our ensemble model to systematically transition its focus, increasingly prioritizing the most recent data, which is presumed to be the most relevant for the current learning context. This approach exemplifies the nuanced and progressive nature of online learning, where the model dynamically adapts to the evolving data landscape, ensuring that its classifications remain accurate and pertinent.

\subsection{Results and Analysis}

Our primary objective is to develop a model adept at classifying data from the current season, which is data from the season 2019-2020 in this study. To achieve this, we trained individual models for each season ranging from 2011 to 2018. Subsequently, we evaluated the performance of these models using all of the data from season 2019-2020 to ascertain their effectiveness in handling present-season data. The results of using these hyperparameters are presented in Table \ref{uniyear}, providing a clear overview of each model's performance across this temporal span.

\begin{table}[h]
\renewcommand{\arraystretch}{1.3}
\caption{Results of individual models}
\centering
\begin{tabular}{|c|c|c|}
\hline
\textbf{Season} & \textbf{AUROC} \\ \hline
2011-2012 & 0.800 \\ \hline
2012-2013 & 0.628 \\ \hline
2013-2014 & 0.637 \\ \hline
2014-2015 & 0.765 \\ \hline
2015-2016 & 0.838 \\ \hline
2016-2017 & 0.867 \\ \hline
2017-2018 & 0.866 \\ \hline
2018-2019 & 0.865 \\ \hline
\end{tabular}
\hspace{0.15cm}
\label{uniyear}
\end{table}

For the results, a clear upward trend in AUROC scores is evident as we approach the season of 2019-2020, although 2011-2012 is an outlier. This trend is especially pronounced in models trained with more recent data, indicating an improvement in the classifiers' ability to distinguish between positive and negative cases as they are trained on data closer to the target season. This improvement can be attributed to the increasing relevance and recency of the data. As the training data gets closer to the target season, it likely becomes more representative of the current patterns and trends, thereby enhancing the classifiers' accuracy.

To evaluate the performance of MSAW strategy, we compare them with baseline methods, including:
\begin{itemize}
    \item Pre-trained Using Source:  This approach involves pre-training a model using a combined dataset from all source domains.
    \item Online Learning Model: This model employs an online learning strategy, focusing exclusively on sequentially acquired data from the target domain for training. In this approach, the model is tested on an incoming data point and then trained on it before moving on to the next, thus continuously updating its parameters in a test-and-train cycle to adapt to new data.
    \item Online Learning Model Pre-trained Using Source:  This approach combines two previous methods together by pre-training a model using all source domains and online learning on the target domain.
    \item Ensemble Method with Equal Weights: This method employs an ensemble method that integrates various individual source models, assigning them equal and fixed weights throughout the process.
    \item Ensemble Method with Volume-based Weights: This method uses an ensemble method that weights individual source models according to the volume of their respective training datasets.
    \item Ensemble Method with Time-based Weights: This ensemble method integrates individual models based on their temporal proximity to the testing season (target domain), defining `time distance' as the number of seasons separating a model's training data from the testing period.
\end{itemize}



In our comparative analysis of various online learning strategies against the backdrop of sequential epidemiological data, distinct performance metrics emerge, highlighting the efficacy and adaptability of these models as shown in Table \ref{ensemble}. 
The model pre-trained using all of the source domains, at an AUROC of 0.852, demonstrates the potential advantages of a comprehensive, cross-domain dataset for pre-training, indicating the value of extensive historical data. The online learning model, with an AUROC of 0.850, showcases a commendable capacity for leveraging sequentially acquired data, underscoring the viability of online learning in dynamically evolving domains. Online learning started with a pre-trained model boosts the AUROC to 0.864. Conversely, the ensemble with equal weights method registers a lower AUROC of 0.728, suggesting that uniform weighting across diverse models may dilute the classification precision. More notably, the ensemble with volume-based weights, at an AUROC of 0.680, indicates that volume-centric model weighting may not effectively capture the intricate data nuances necessary for accurate target domain classifications. An improvement is observed with the ensemble with time-based weights, achieving an AUROC of 0.761, which implies that temporal relevance plays a crucial role in enhancing model performance, albeit not to the extent of the more sophisticated methods. 
Our MSAW strategy exhibits an AUROC of 0.870, which is superior to all other methods. This delineates the MSAW strategy's predominant advantage in dynamically calibrating model weights based on the evolving data landscape, marking a significant stride towards optimizing predictive models for healthcare applications, particularly in the nuanced domain of RSV data analysis.



\begin{table}[h]
\renewcommand{\arraystretch}{1.3}\caption{Comparative results of different models, featuring P values from DeLong's test for statistical differences between one AUROC and the AUROC of the MSAW method.}
\centering
\begin{tabular}{|l|l|}
\hline
\textbf{Method} & \textbf{AUROC}\\ \hline
Pre-trained Using Source & 0.852 (P$<$0.001) \\ \hline
Online Learning & 0.850 (P=0.037) \\ \hline
Online Learning Model Pre-trained Using Source & 0.864 (P$<$0.001)  \\ \hline
Ensemble with Equal Weights & 0.728 (P$<$0.001) \\ \hline
Ensemble with Volume-based Weights & 0.680 (P$<$0.001) \\ \hline
Ensemble with Time-based Weights & 0.761 (P$<$0.001)  \\ \hline
MSAW & 0.870 \\ \hline
\end{tabular}
\label{ensemble}
\end{table}



\begin{figure*}[h]
    \centering
    \includegraphics[width=170mm]{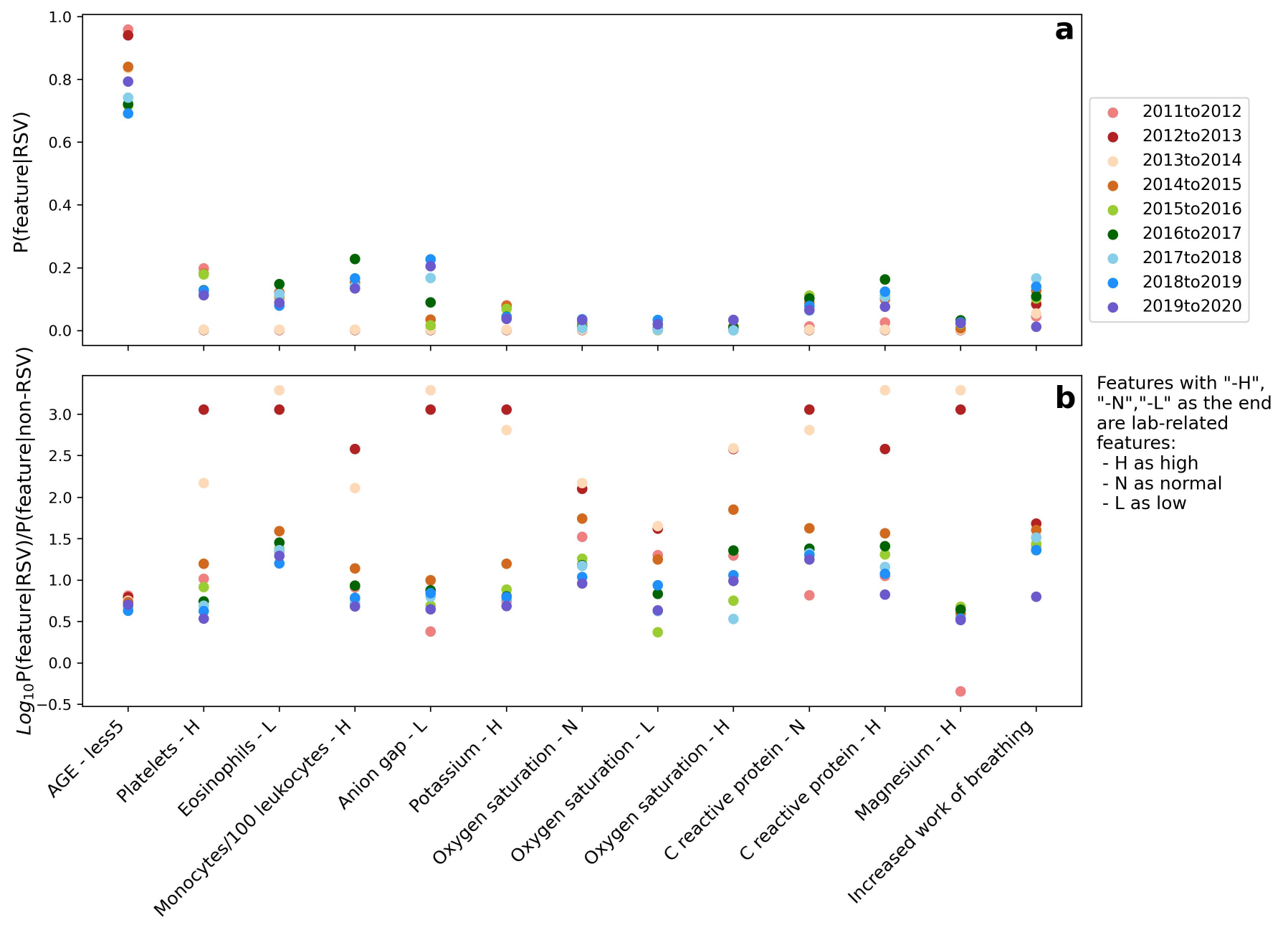}
    \caption{ (a) Probability of a feature present given RSV \\
    (b) $Log_{10} LR$ of features present.}
    \label{fig:possibility}
\end{figure*}



Our proposed MSAW approach, which utilizes dynamically updated weights, surpasses these conventional baseline methods. The essence of its enhanced performance lies in its ability to dynamically adapt, customizing its relevance to the current data. This dynamic adaptation is achieved by continuously adjusting the weights of the models based on their immediate performance with new data. Such a strategy enables the model to respond effectively to changing data trends and to identify patterns that were not apparent in earlier training datasets. As a result, our method shows an improvement in both accuracy and effectiveness in classifying medical data, outperforming the other methodologies evaluated in the study.

Fig. \ref{fig:possibility}(a) presents a comparison of the probabilities of features present in RSV cases across different seasons, as calculated in our training dataset. Fig. \ref{fig:possibility}(b) compares the $Log_{10}$ likelihood ratios (LR) between RSV cases and non-RSV encounters. A positive LR value suggests that the presence of the feature is more probable in RSV cases than in non-RSV encounters. Among the features for which the probability P(feature$|$RSV) exceeded 0.01 in the 2019-2020 season, we identified the top 13 features with the LR. In Fig. \ref{fig:possibility}(a), the probability of `Age-less than 5' group appears high with values ranging from 0.68 to 0.96 in all years, while other features only have around or below 0.2. As shown in Fig. \ref{fig:possibility}(b), with relatively small ranges of values, `Age-less than 5' and `Increased work of breathing' are two significantly steadier features than others observed through multiple seasons. The values in models of seasons 2012-2013 and 2013-2014 are the most divergent from the model of 2019-2020 season, suggesting that the feature patterns in the 2012-2013 and 2013-2014 seasons are the least similar to our target season in 2019-2020, which also aligns with 
our model results in Table \ref{uniyear} where seasons 2012-2013 and 2013-2014 present the lowest AUROC. Conversely, the values for models corresponding to the 2017-2018 and 2018-2019 seasons are predominantly found nearest to the 2019-2020 season, and their respective models also demonstrate strong performance, as detailed in Table \ref{uniyear}.

\section{Discussion}


In this study, we have analyzed various ensemble methods applied to the challenging task of medical data classification. The primary goal was to develop a robust classifier for the season of 2019-2020, leveraging historical data spanning from 2011 to 2018. The results obtained from our experiments do not necessarily reveal a temporal trend in the performance of individual models.
The model developed using data from season 2011-2012 performed much better than season 2012-2013, 2013-2014, and 2014-2015, highlighting the importance of allowing the model to autonomously discern relevant patterns and relationships. This finding emphasizes the importance of adopting dynamic and data-driven approaches in medical data classification, as the temporal relevance of information may not always align with conventional intuitions.

The evaluation of ensemble methods against baseline models provided valuable insights. Although the AUROCs of the  `Online Learning Model Pre-trained Using Source' method and our proposed MSAW are comparable, the MSAW is performing slightly better and is more adaptable to situations in which recent data experience a major domain shift. The MSAW might be expected to perform significantly better when there is a large domain shift in the recent past, for example, new strains of RSV that are symptomatically quite different than previous strains. The MSAW method outperformed other techniques. This superior performance can be attributed to the method's inherent adaptability, which allows it to continuously learn and assign weights to individual models based on their performance on the most recent data. By doing so, our approach effectively captures emerging patterns and evolving relationships within the data, resulting in enhanced classification accuracy. In an environment where medical knowledge and practice are continually evolving, our research underscores the potential of dynamic ensemble methods to provide more accurate and efficient solutions for medical classification tasks. These findings have significant implications for healthcare decision-making, as they demonstrate the importance of data-driven adaptability in capturing the intricate nuances of medical data. 

As shown in Fig. \ref{fig:possibility}, many laboratory tests are correlated with RSV presence. For instance, there is a notable correlation between higher platelet counts and the presence of RSV. This correlation is also mentioned in \cite{elber2023platelet}, which has shown that children diagnosed with RSV tend to have significantly higher platelet counts compared to those with negative respiratory virus panels or infections caused by other respiratory viruses, including HMPV, influenza A, influenza B, Adenovirus, and Parainfluenza. 
As another example, our models indicate that patients with RSV often exhibit higher Monocytes counts per 100 leukocytes. One research \cite{kimpen2001respiratory} aligns with our observation, showing that RSV may trigger monocyte activation. Interestingly, additional findings reveal that even normal levels of oxygen saturation and C-reactive protein are associated with RSV. The act of performing these tests, as opposed to not performing these tests, shows a relationship with RSV presence. 
All these findings highlight the effectiveness of using observational EHRs to detect correlations between lab results and RSV. Importantly, Fig. \ref{fig:possibility} demonstrates annual variations in the correlation strength between laboratory results and RSV, possibly due in part to adjustments in laboratory testing protocols for different age groups and patient severity levels.

In spite of the promising results presented in this study, there remain several noteworthy limitations that warrant attention in future research. In our study, while the Naive Bayes models proved effective, they represent only one simple modeling technique for medical data classification. For more intricate classification tasks, the Naive Bayes method may not effectively capture the complexities of medical data as well as more advanced algorithms, such as neural networks. Therefore, future research should explore incorporating these advanced machine learning techniques, which may offer an improvement in the modeling of the nuanced patterns in medical data. Additionally, the dynamic penalty factor used in our study, though effective, has room for refinement. For example, enhancing its adaptability to the changing data distributions in the target domain could lead to more accurate and responsive models. Specifically, rather than uniformly penalizing all weights, we aim to explore more sophisticated methods to adjust the weights of the model's parameters for different features to varying extents. This approach would enable a more accurate reflection of changes.

Another limitation of our study stems from our dataset. RSV cases are identified using ICD codes rather than confirmed through RSV testing results. Consequently, non-RSV encounters might include some RSV-positive cases lacking RSV-related ICD codes. However, given the vast number of non-RSV encounters, this issue is unlikely to significantly impact the results. In addition, the specificity of our dataset, which was centered around RSV information from UPMC, presents a potential constraint in terms of the generalizability of our findings across different medical contexts. The healthcare field is diverse, and patterns observed in one dataset may not be universally applicable. Our future research will aim to validate the model across a broader spectrum of biomedical scenarios. For example, we can incorporate data from different medical institutions, which would help assess the transferability of our approach and determine necessary modifications for different settings. Expanding the scope of our research through collaborative and cross-disciplinary studies could also provide insights and data sources, enhancing the model's applicability and robustness.

\section{Conclusion}
This paper introduces an online multi-source transfer learning methodology featuring a dynamic weighting mechanism, applied to the development of RSV case detection models using multiple years' EHR data from the UPMC. Experimental results demonstrate that this method significantly outperforms many baseline techniques, highlighting the considerable potential of the approach. Notably, it adeptly addresses the challenges inherent in biomedical research, such as the scarcity of labeled data and the complexities associated with sequential data. The results of our investigation provide support for the effectiveness of our MSAW method, but also pave the way for studying more advanced classification models for sequential data analysis.

\section{Acknowledgements}
This work was supported by the research grants
R00LM013383 and R01LM013509 from the National Library of Medicine, National Institutes of Health (NIH). The research data were retrieved by
the University of Pittsburgh's Health Record Research Request, which was supported by the NIH grant UL1TR001857. 
Dr. Harry Hochheiser was supported by MIDAS grant U24GM132013. Dr. Jessi Espino was partially supported by the National Center for Immunization and Respiratory Diseases grant 5U01IP001184.
We are grateful for the invaluable assistance and guidance from Dr. Richard K. Zimmerman and Dr. Mary Patricia Nowalk, and Dr. Milos Hauskrecht, from the University of Pittsburgh.

\bibliographystyle{IEEEtran}
\bibliography{ref}

\end{document}